\documentclass[pmlr]{jmlr2}% new name PMLR (Proceedings of Machine Learning Research)
\usepackage{longtable}% for long tables

\usepackage{booktabs}
\usepackage[load-configurations=version-1]{siunitx} % newer version
\usepackage{dsfont}
\usepackage[nolist,nohyperlinks]{acronym}
\usepackage{ulem} % \sout{}, strike-through text
\usepackage{xcolor}    
\usepackage{wrapfig}
\usepackage{graphicx}
\usepackage{float}
\usepackage{caption}
\begin{acronym}
\acro{ML}{{Machine Learning}}
\acro{UQ}{{Uncertainty Quantification}}
\acro{CP}{{Conformal Prediction}}
\acro{CRC}{{Conformal Risk Control}}
\acro{AI}{{Artificial Intelligence}}
\acro{OD}{{Object Detection}}
\acro{TLL}{Traffic Light Localization}
\end{acronym}
\newcommand{\R}{\mathbb{R}}
\newcommand{\ml}{\ac{ML}\xspace}
\newcommand{\uq}{\ac{UQ}\xspace}
\newcommand{\cp}{\ac{CP}\xspace}
\newcommand{\ai}{\ac{AI}\xspace}
\newcommand{\od}{\ac{OD}\xspace}

\newcommand{\crc}{\ac{CRC}\xspace}
\newcommand{\tll}{\ac{TLL}\xspace}

\newcommand{\yolov}[1]{YOLOv5{#1}}
\newcommand{\ddet}{DiffusionDet\xspace}
\newcommand{\detr}{DETR-ResNet50\xspace}

\newcommand{\ccol}[1]{\multicolumn{1}{c}{#1}} % shortcut to CENTER THE COLUMN NAME while using l/c/r on the tabular specification

\newcommand{\dx}{ \widehat{w}^{k}}
\newcommand{\dy}{ \widehat{h}^{k}}
\newcommand{\w}{ \widehat{w}}
\newcommand{\h}{ \widehat{h}}

\definecolor{greenish}{RGB}{55, 157, 143}

\def\E{\mathbb{E}} %% EXPECTED VALUE
\newcommand{\C}{\mathcal{C}} %% C for set: C(X)
\renewcommand{\P}{\mathbb{P}} %% P of probability
\renewcommand{\max}{\text{max}} %% P of probability
\newcommand{\f}{ \widehat{f}}
\newcommand{\wx}{ \widehat{x}}
\newcommand{\wy}{ \widehat{y}}

\newcommand{\dtrain}{D_{\text{train}}}
\newcommand{\ntrain}{n_{\text{train}}}

\newcommand{\dfit}{D_{\text{fit}}}
\newcommand{\dcal}{D_{\text{cal}}}
\newcommand{\ncal}{n_{\text{cal}}}
\newcommand{\dtest}{D_{\text{test}}}

\newcommand{\xnew}{X_{n+1}}
\newcommand{\ynew}{Y_{\text{new}}}

\newcommand{\cupdot}{\mathbin{\mathaccent\cdot\cup}}

\newcommand{\ie}{i.e.\xspace}
\newcommand{\eg}{e.g.\xspace}

\jmlrvolume{}%1}
\jmlryear{}%2023}
\jmlrworkshop{}

\title[Conformal Object Detection]{Confident Object Detection via Conformal Prediction and Conformal Risk Control: an Application to Railway Signaling}

\author{\Name{L\'{e}o And\'{e}ol}  \Email{leo.andeol@math.univ-toulouse.fr} \\
        \addr Institut de Math\'{e}matiques de Toulouse, Toulouse, France\\
        \addr SNCF, Saint-Denis, France
    \AND
    \Name{Thomas Fel} \Email{thomas\_fel@brown.edu} \\
        \addr Brown University, Providence, Rhode Island, USA\\
        \addr SNCF, Saint-Denis, France
    \AND 
    \Name{Florence {de Grancey}} \Email{florence.de-grancey@irt-saintexupery.com} \\
        \addr Thales AVS France SAS 
    \AND 
    \Name{Luca Mossina} \Email{luca.mossina@irt-saintexupery.com} \\
        \addr IRT Saint Exup\'{e}ry, Toulouse, France \\
}

\editor{Editor's name}

\begin{document}

\maketitle

\begin{abstract}

Deploying deep learning models in real-world certified systems requires the ability to provide confidence estimates that accurately reflect their uncertainty. In this paper, we demonstrate the use of the conformal prediction framework to construct reliable and trustworthy predictors for detecting railway signals. Our approach is based on a novel dataset that includes images taken from the perspective of a train operator and state-of-the-art object detectors. We test several conformal approaches and introduce a new method based on conformal risk control. Our findings demonstrate the potential of the conformal prediction framework to evaluate model performance and provide practical guidance for achieving formally guaranteed uncertainty bounds.
\end{abstract}


\vspace{-2mm}
\section{Introduction} \label{sec:intro}

The deployment of \ml technologies in real-world, safety-critical systems is faced with many challenges;
one of them is to provide \uq for the output of the ML component.
While this quantification can be accessible for low-complexity models, this is an important challenge for complex tasks such as object detection or text processing.

In this paper we explore how Conformal Prediction \citep[CP]{vovk_2022_alrw} and Conformal Risk Control \citep[CRC]{angelopoulos_2022_conformal_risk} can contribute to build confident (or ``trustworthy'') predictors for the task of \od \citep{zhao_2019_object}.

CP and CRC have the advantage of being distribution-free, non-asymptotic and model-agnostic frameworks,
which allow their deployment to any black-box predictor under minimal hypotheses, including complex ML tasks. 

Furthermore, they are computationally lightweight as they do not require retraining the model, and so can easily be added to existing \ml pipelines\footnote{This holds true for \textit{split} (or ``inductive'') \cp, which is the only form of \cp we use for our applications.}.

In this work, we demonstrate on a purpose-built dataset how the combination of these frameworks with state-of-the-art object detection models can lead to more accurate and reliable predictions in real-world applications.

After introducing our use case in Section~\ref{sec:use-case}, we detail the construction of our railway signaling dataset in Section \ref{sec:dataset-construction}. Then, in Section~\ref{sec:uq-od} we provide an overview of \cp, \crc and some related methods. % in \uq for \od.
In Section~\ref{sec:build-cp-od} we give the details of our approach. Afterwards, in Section~\ref{sec:experiments} we set up our experiments and discuss some important methodological details.

Finally, in Section~\ref{sec:results} and ~\ref{sec:conclusion} we discuss the results, conclude on our work and give some insights, as well as leads for future works.

\subsection{Railway Traffic Light Detection} \label{sec:use-case}

\begin{figure}[t]
    \centering
    \includegraphics[width=0.50\textwidth]{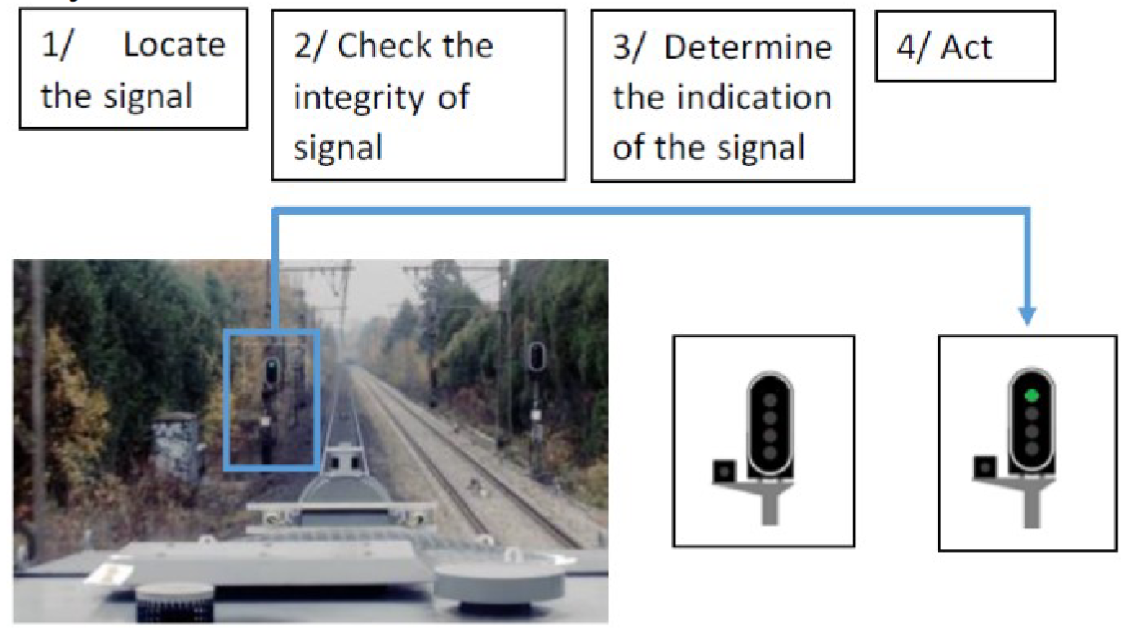}
    \caption{ 
    Example of a pipeline where an \ai system acts following \ml-based predictions. Source: \citet{alecu_2022_can}.}
    \label{fig:train-od-pipeline}
\vspace{-4mm}
\end{figure}

Our use case consists in detecting
the signals as they are encountered during the operation of trains in a railway network, and we refer to this problem as \textit{Railway Traffic Light Detection} (RTLD).
 
While main lines (\eg high-speed lines) already have in-cabin signaling and can be automatized \citep{singh_2021_deployment}, this is too costly to be applied to the whole network.
Consequently, on secondary lines, drivers can be subject to a larger cognitive load to interpret signals and the environment. 
Assisting drivers with \ai-based signaling recognition could facilitate the operations.  
This operational {desideratum} can be cast as a functional chain including: locating the traffic light, validating the localization and recognizing the signal class. %, as, presented in  
In \citet{alecu_2022_can}, who provide an overview of the technical and regulatory challenges raised by the safety of \ai systems in the railway and automotive industries,

we find a study of this problem as depicted in Figure~\ref{fig:train-od-pipeline}.
Our application (referred to as \textit{Traffic Light Localization}, TLL) would correspond to point (1).

\subsection{Theoretical guarantees in ML with conformal prediction}

Providing \uq for the output of an object detector can be interpreted as providing a set of points (\eg pixels) that is likely to contain the (pixels of the) ground-truth box.
Concretely, they can be computed by adding a ``safety'' margin around the sides of the predicted bounding boxes, according to some statistical property specified by the user.
This could be the average \textit{coverage} of our procedure, that is, the frequency with which we capture the ground truth during inference.
Letting $X_{\text{new}}$ be the observed features of a new sample, we want a set predictor $\C(X_{\text{new}})$ that contains \textit{entirely} the ground-truth box $Y_{\text{new}}$ with probability:

\begin{equation}
    \P\Big( Y_{\text{new}} \in \C(X_{\text{new}}) \Big) \geq 1 - \alpha,
    \label{eq:cp-coverage}
\end{equation}

Equation~\ref{eq:cp-coverage} represents the core theoretical guarantee of the conformal prediction framework, which we will employ to analyze the outputs of our object detector. 
During inference, a conformal algorithm constructs a prediction set $\C_{\alpha}(X)$ that, on average, covers the observed value of the target $Y$ with a frequency of $1-\alpha$ across multiple repetitions of the procedure. 

The challenge is to formulate the problem and build a set predictor $\C(\cdot)$ that answers to an operational need, such as ``capturing the entire bounding boxes $(1-\alpha) ~ 100\%$ of the time'' (for \cp) or  ``covering at least $(1-\alpha) ~ 100\%$ of the target pixels'' (for \crc, Section~\ref{sec:build-cp-od}).

\section{Building an experimental dataset for object detection}
\label{sec:dataset-construction}
We work on the detection of traffic lights in the French railway network.
Since most national railway networks have a unique mix of signals and traffic lights, one needs to build a dataset for the specific operational domain. %, as detailed below.

\subsection{Related work}
Public datasets for computer vision on railway data are rather scarce. 
\citet{Zendel_2019_RailSem19} built the first public, railway-specific dataset for semantic scene understanding, which includes the manual annotation of geometric shapes and pixel-wise labeling.
They also provide an overview of publicly available datasets containing in part railway data.
To counter this scarcity,
\citet{Gasparini_2020_anomaly} collected 30k night-time images, captured by a drone flying over the rails, for the task of detecting autonomously anomalous objects on rails.
Another viable option is to use artificial data, as done for instance by \citet{Mauri_2022_Lightweight}, who created a virtual dataset from a simulator based on a video game.
For the French network, \citet{Zouaoui_2022_railset} created a segmentation dataset with artificially generated anomalies.
The dataset \textit{FRSign} of \citet{harb_2020_frsign} addresses a similar use case via object detection: they were able to coordinate the data collection with the national railway operator and other partners.

In our case, we found an alternative sourcing existing videos from the internet.
Also, compared to the work of \citeauthor{harb_2020_frsign}, our new dataset presents increased variability (more railway lines, environmental and weather conditions, etc.) which could enable more accurate predictions in real-world scenarios.
Finally, we generalize the task from single to multi-object detection, laying the foundations for future work in instance segmentation.

\subsection{Dataset characteristics}
 
The \tll should operate whenever a human operator typically operates.
This leads to considering a high-diversity dataset, including various meteorological situations (rain, snow, etc.), various hours (night, day) and a wide variety of traffic light situations (fully observable, partially occluded by foliage, etc.).
To partially account for these common issues, we included different light conditions in our data.

\begin{table}[h] 
    \centering
    \small
    \begin{tabular}{l r}
        Characteristics & \multicolumn{1}{c}{Quantity} \\
        \midrule
        Railway lines               & 41 \\
        Images per line (average)   & 83.27 $\pm$ 41.11 \\
        Images in dataset           & 3414 \\
        {Dimensions} (pixels)       & 1280 $\times$ 720  \\
        Bounding boxes per image      & 1.03 $\pm$ 1.26 \\
        Bounding boxes (total)      & 3508 \\
    \end{tabular}
    \caption{Characteristics of our dataset} 
    \label{tab:dataset-composition}
\end{table}

As source data, we used footage from 41 videos of French railway lines, freely available on the internet, with the approval of the uploader\footnote{We would like to thank the author of the Youtube channel: \url{https://www.youtube.com/@mika67407}}. Most of the railway lines are distinct, but a few share sections especially around large Parisian stations. The average duration of a video is about 1.5 hours, from which we'll extract individual frames. 
The extraction of frames was conducted as follows:
We extract frames from videos by running a pretrained object detector with a low objectness threshold, and we keep a minimum interval of 5 seconds between detections so as to avoid different frames featuring the same signals, to prevent excessive temporal correlation between images. On average, 83 frames were extracted per video. We then keep only the frames, and {manually} and individually annotated all visible railway signals on them.
In Table~\ref{tab:dataset-composition} we report the statistics of our dataset.

\section{Uncertainty quantification in object detection} \label{sec:uq-od}

For our tests, we restricted our attention to \yolov{m} \citep{Jocher_2020_YOLOv5}, originally proposed by \citep{redmon2016you}, \detr \citep{Carion_2020_detr} and \ddet \citep{chen2022diffusiondet}.
\yolov{m} offers a one-stage detection, combining convolutional layers with regression and classification tasks, and has found widespread adoption.
\detr leverages transformer layers and \ddet formulated \od as a denoising diffusion problem.
These were chosen because they are either standard models, or state-of-the-art ones.
Since \cp and \crc are model-agnostic, the choice of \od network does not matter. 
For instance, \citet{Petrovic_2022_integration} build a detector of railway tracks and signals.

Also, the application of \cp is not limited to traffic lights (our use case) but can be extended to any detection needing formal guarantees \citep{Ye_2020_autonomous}.
Applying \cp to specialized models could open up future lines of research.

\subsection{Related works}
In industrial applications,  
it is often hard to make reliable hypotheses on the data or the correct specification of the predictor, which is why we favored a distribution-free, model-agnostic approach such as conformal prediction.
Of course, if one {can} make meaningful assumptions on their learning task, then there is a sizeable literature on the topic 
\citep[for a review]{feng2021review}. 
\citet{hall2020probabilistic} give a probabilistic formulation of \od,
where the probability distributions of the bounding boxes and classes are predicted.
Bayesian models like in \citet{harakeh2020bayesod} and Bayesian approximations \citep{deepshikha2021monte} are also found in the literature. 
We point out the distribution-free approach of \citet{li_2022_towards}: they build probably approximately correct prediction sets using a held-out calibration set to compute a calibrated threshold for the predictor (e.g. for the softmax), following the principles introduced by \citet{Park_2020_PAC_sets}.
They control the coordinates of the boxes but also the proposal and objectness scores, resulting in more and larger boxes. 
Their method relies on the structure of Fast R-CNN \citep{Ren_2017_Faster_R_CNN}, the underlying \od model: this has three detection steps with three predictors associated with the \textit{proposal}, \textit{presence} and \textit{location} of a bounding box.
Each component is controlled individually and then combined to attain the desired
guarantee.
Their method is an application of the PAC-based calibration of \citet{Park_2020_PAC_sets}.
This is not applicable \textit{as is} to state-of-the-art one-stage object detectors such as \textit{YOLO} \citep{redmon2016you} or \textit{DETR} \citep{Carion_2020_detr}.
This is one of the reasons why we opt to model our uncertainty quantification problem directly via \cp.
Also, \cp requires exchangeable data while concentration-based methods such as \citet{Park_2020_PAC_sets} and the more general methods of \citet{bates_2021_rcps} and \citet{angelopoulos_2022_LTT}require the stronger assumption of data being independently and identically distributed (i.i.d).

\subsection{Principles of Conformal Prediction} \label{sec:principles-cp}

\acf{CP} \citep{vovk_2022_alrw, Angelopoulos_2023_Gentle} 
is a family of methods to perform \uq with guarantees under the sole hypothesis of data being independent and identically distributed (or more generally exchangeable). 
\cp is flexible because we can either ``conformalize''\footnote{
    We (loosely) say that we ``conformalize'' a predictor $\f$ whenever we apply either \cp or \crc. For \cp, a \textit{conformalized} model is one that outputs a prediction set (\eg enlarged bounding box) that does not contain the target $Y$ at most with frequency $\alpha$.
} a predictor using the training data, for instance via {transductive} ``full'' \cp \citep{vovk_2022_alrw} or a $K$-fold partition scheme \citep{Vovk_2013_cross-conformal, Barber_2021_jp},
or via a dedicated calibration dataset $\dcal$ with a method known as \textbf{split} \cp \citep{papadopoulos_2002_inductive, lei_2018_distribution}.
This allows using a \textbf{pretrained} predictor $\f$ with no need to access the training data.
Throughout the paper, ``CP'' always refers to Split \cp; % and limit our exposition to this case.
unless otherwise specified, we write $n = |\dcal| = \ncal$ and $(X_{n+1},Y_{n+1})$ will denote a (random) test point drawn from the same distribution as $\dcal = \{(X_i, Y_i)\}_{i-1}^{n}, (X_i,Y_i) \sim \P_{XY}$.

For a specified (small) error rate $\alpha \in (0,1)$ and $n$ calibration points, during inference, the \cp procedure will yield a prediction set $C_{\alpha}(X_{n+1})$ that fails to cover the observed $Y_{n+1}$ with probability at most $\alpha$: 

\begin{equation}
    \P\Big( Y_{n+1} \notin \C_{\alpha}(X_{n+1}) \Big) \leq \alpha.
    \label{eq:cp-miscoverage}
\end{equation}

Formally, this guarantee holds true, on average, 
over many repetitions of the \cp procedure (\ie sampling of calibration and test points).
It is valid for any distribution $\P_{XY}$, any sample size and any predictive model $\f$, even if it is misspecified or a black box.
The probability $1-\alpha$ is referred to as the \textit{nominal coverage};
the \textit{empirical coverage} on $n_{\text{test}}$ test points is $\frac{1}{n_{\text{test}}} \sum_{i=1}^{n_{\text{test}}} \mathds{1}\{Y_i \in {C}_{\alpha}(X_i)\}$.

The conformalization of $\f$ is determined by a \textbf{nonconformity score} $s(X, Y)$, measuring how ``unusual'' the prediction $\widehat{Y} = \f(X)$ is with respect to observed $Y$.
This is a generic method: 
\eg, for regression 
we can set $s(X,Y) = |\widehat{f}(X) - Y|$;
for quantile regression \citep{Koenker_1978_quantile},
we can measure the errors of lower and upper quantile estimators $(\widehat{q}_{\beta}, \widehat{q}_{1- \beta})$ with $s(X,Y) = \max \{\widehat{q}_{\beta}(X) - Y, Y - \widehat{q}_{1 - \beta}(X)  \}$ of \citet{Romano_2019_CQR}.

\begin{algorithm2e} 
\caption{Split conformal prediction: \textit{fit}, \textit{conformalization} and \textit{inference} steps.}
\label{alg:vanilla-split-cp}
 \DontPrintSemicolon 
 %\LinesNumbered
 \KwIn{Training data $\dtrain = \{(X_i, Y_i)\}_{i=1}^{\ntrain}$; miscoverage level $\alpha \in (0,1)$; nonconformity score $s(X,Y)$.}
    \begin{enumerate} \itemsep0em
        \item Split (disjointly) training data: $\dtrain = \dfit  \cupdot \dcal$ 
        \item Fit (or fine-tune) $\widehat{f}(\cdot)$ on $\dfit$ 
        \item Compute scores on $\dcal$: $\bar{R} = \{ s(X_i, Y_i) \}_{i=1}^{\ncal}$
        \item Compute conformal quantile: $q_{1 - \alpha} = \lceil (n_{\text{cal}}+1)(1-{\alpha}) \rceil$-th element of the {\it sorted} sequence $\bar{R}$
        \item Inference: $\C_{\alpha}(X_i) = \Big\{ y: s(X_i, y) \leq q_{1 - \alpha} \Big\}.$ \label{eq:split-cp-set} 
    \end{enumerate}
\end{algorithm2e}

In Algorithm~\ref{alg:vanilla-split-cp} we give the steps of Split \cp.
If one uses a pretrained predictor, then only $D_{\text{cal}}$ are needed and Steps 1 and 2 are skipped.
During {conformalization}, we compute the {nonconformity scores}  $\bar{R}$ on $\dcal$.  
For a test point $\xnew$, we build the prediction set $\C_{\alpha}(\xnew) = \{ y: s(\xnew, y) \leq q_{1 - \alpha} \}.$ 
For example, if $s(X,Y) = |Y - \widehat{f}(X)|$, then we build the prediction interval as  
$\C_{\alpha}(\xnew) = [\widehat{Y} - q_{1 - \alpha}, \widehat{Y} + q_{1 - \alpha} ].$
In Section~\ref{sec:build-cp-od} we show that, for our \tll case, \cp boils down to adding a margin around the predicted bounding boxes.

\subsection{Conformal risk control: a generalization of conformal prediction}
\citet{angelopoulos_2022_conformal_risk} introduced \acf{CRC} as an extension of \cp.
First, they point out that the conformal guarantee in Equation~\ref{eq:cp-miscoverage} can be rewritten as $\E [ \mathds{1}{\{ \ynew \notin \C_{\alpha}(\xnew) \}} ] \leq \alpha$.
The function $\ell \big( \C_{\alpha}(\xnew), \ynew \big) = \mathds{1}{\{ \ynew \notin \C_{\alpha}(\xnew) \}}$ encapsulates a \textit{notion of error},  
which for \cp occurs whenever $\ynew$ is not covered by $\C_{\alpha}(X)$.
In some practical applications, this binary loss can be too strict and building a prediction set according to another criterion can satisfy (theoretically) a different operational need (\eg false negative rate).
The \cp procedure can be extended to {any bounded loss function $\ell(\cdot)$}, provided that it decreases as the set $\C(X_{n+1})$ gets larger;
the task is generalized as $\E\big[ \ell \big( \C(X_{n+1}), Y_{n+1} \big) \big] \leq \alpha$, where $\C(X_{n+1})$ is not necessarily built by \cp. 
Let $\f(X)$ be a pretrained predictor and $\C_{\lambda}(X; \f)$ a function parametrized by $\lambda$, where larger $\lambda$ values yield larger prediction sets.
Given a calibration dataset $\dcal = (X_i, Y_i)_{i=1}^{n}$,
\crc boils down to computing the losses $L_{i}(\lambda) = \ell \big(\C_{\lambda}(X_i), Y_i \big) \in (- \infty, B], B < \infty$ on $\dcal$, 
the empirical risk $\widehat{R}_{n}(\lambda) = \frac{1}{n} (L_1(\lambda) + \dots + L_n(\lambda) )$
and choosing a $\widehat{\lambda}$ such that the risk on the $(n+1)$-th (unseen) sample is controlled: 

\begin{equation}
    % \label{eq:crc-guarantee}
    \label{eq:crc-loss-control}
    \E\Big[ \ell \big(\C_{\hat{\lambda}}(X_{n+1}), Y_{n+1} \big) \Big] \leq \alpha.
\end{equation}

For an arbitrary risk level upper bound $\alpha \in (-\infty, B]$, $\widehat{\lambda}$ is computed as:

\begin{equation}
    \widehat{\lambda} :=\, \text{inf} \Big\{ \lambda: \frac{n}{n+1}\widehat{R}_{n}(\lambda) + \frac{B}{n+1} \leq \alpha \Big\}.
    \label{eq:crc-lambda-estimation}
\end{equation}

Here, $\C_{\hat{\lambda}}$ denotes any set predictor that complies with an $\alpha$ risk level, not necessarily a probability.
Throughout the paper, however, we have losses $L(\lambda) \in [0,1]$ and $\alpha \in (0,1)$.

\subsubsection{CRC covers the conformal prediction case} 
If we consider a miscoverage loss 
$L_i^{\text{coverage}}(\lambda) = \mathds{1}\big\{Y_i \not \in \widehat{\C}_{\lambda}(X_i) \big\} = \mathds{1}\big\{ s(X_{i}, Y_i) > {\lambda}  \big\}$, 
\citet{angelopoulos_2022_conformal_risk} shows that \crc finds the same prediction set as \cp, for a given $\alpha$. 
We can write the CRC prediction set as:

\begin{equation}
    \C_{\widehat{\lambda}}(X_{n+1}) = \{ y: s(X_{n+1}, y) \leq \hat{\lambda} \},
\end{equation}

\noindent where $\hat{\lambda}$ is the same as the conformal quantile $q_{1-\alpha}$ 
of Line~\ref{eq:split-cp-set} of Algorithm~\ref{alg:vanilla-split-cp} for Split \cp. 
The \crc guarantee is less tight than the one proved by \citet{lei_2018_distribution} for Split \cp, 
the latter being $1 - \alpha \leq \P( Y \in C_{\alpha}(X)) \leq 1 - \alpha + \frac{1}{n+1}$ while the former is $1 - \alpha \leq \P( Y \in C_{\hat{\lambda}}(X)) \leq 1 - \alpha + \frac{2 B}{n+1}$.

\section{Building Conformal Predictors for Object Detection} \label{sec:build-cp-od}
In our experiments, we test \acf{CP} and \acf{CRC} in \od. 
For the first part, we follow the box-wise \cp methods of \citet{degrancey_2022_detection} and \citet{andeol_2023_conformal}.
To the best of our knowledge, these are the only straightforward applications of \cp to object detection. 
In addition to their methods, we also test the new, better-performing, max-additive and max-multiplicative scores (see Section~\ref{sec:cp-od-scores}).
For the second part of our experiments, we compare the image-wise \cp method of \citet{degrancey_2022_detection}, to our approach which relies on the \acf{CRC} of \citet{angelopoulos_2022_conformal_risk}, who extend \cp to a more general class of errors: while \cp provides a guarantee on a binary error ``the truth is \textit{contained} vs \textit{not contained} in the prediction set'', \crc admits more generic guarantees of the type ``$(1-\alpha)
 ~100\%$ of the pixels will be covered by the CRC output'' (see Section~\ref{sec:crc}).

We define objects we work with as follows: the output of the \od predictor is a variable-sized (potentially empty) set of bounding boxes $\f(\xnew) = \widehat{Y}_{n+1} = \{ \widehat{Y}_{n+1}^k \}_{k=1, \dots, n_{\text{i}}}$. However, unlike previous work of \citet{degrancey_2022_detection} which considers conjunctions of half-spaces, we will note our bounding boxes in the common \od standard, as a set of four coordinates $\{\wx_{\text{min}}^{k}, \wy_{\text{min}}^{k} ,  \wx_{\text{max}}^{k}, \wy_{\text{max}}^{k} \}$ to ensure wider understanding among the \od community.
However, this notation is imprecise, and therefore we also adopt an implicit definition of bounding boxes as the set of pixels that belong to them, which is closely related to segmentation and necessary for the proper definition of some of our proposed methods. 
Each box $\widehat{Y}_i^k$ is therefore the set of pixels
\[
\widehat{Y}_{i}^{k} = \Bigg\{ (x,y) \in \mathbb{R}^{2} :
\begin{array}{l}
x \in [~\wx_{\text{min}}^{k}~,~ \wx_{\text{max}}^{k}\hspace{0.3mm}] \\
y \in [~\wy_{\text{min}}^{k}~,~ \wy_{\text{max}}^{k}~] \\
\end{array}
\Bigg\}.
\]
In all cases, ground truth boxes are defined equivalently. Based on these definitions, we can introduce two different approaches to guaranteeing \od predictions, box-wise or image-wise guarantees.

\begin{wrapfigure}{r}{0.5\textwidth}
\vspace{-10mm}
\center
\includegraphics[width=0.5\textwidth]{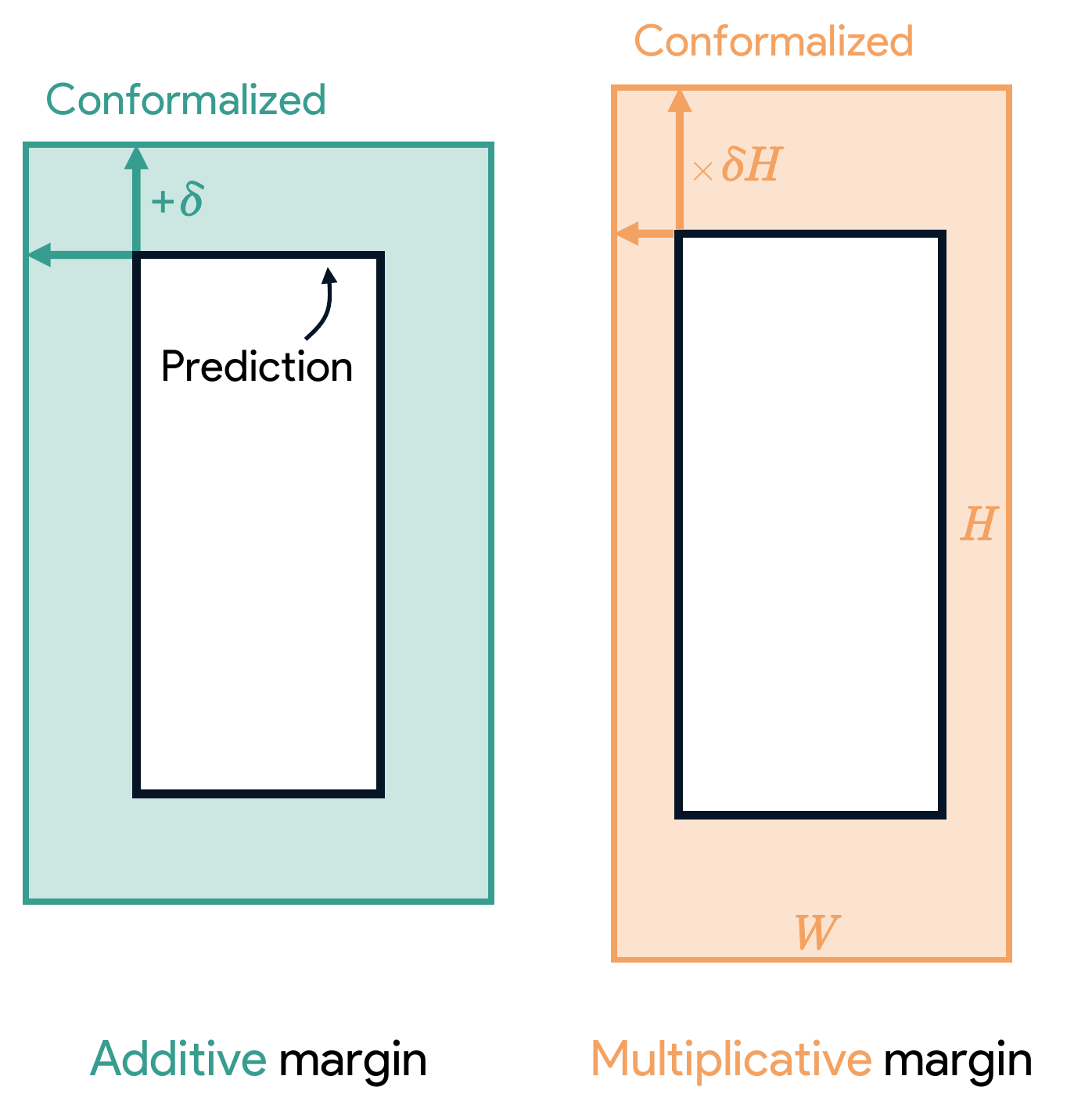}
\caption{\textbf{Effect of margin systems used.}
 An additive margin is a number of pixels to add, while a multiplicative margin is a proportion of the width/height to add. The additive one may lead to comparatively a smaller effect on foreground boxes and larger on background (smaller traffic signals) boxes, and the opposite applies for multiplicative margins.
 }
\label{fig:margins}
\vspace{-5mm}
\end{wrapfigure}

\subsection{Box-wise Conformalization}

The most intuitive approach to conformalize object detector predictions is to work box-wise, that is to to consider our $Y_i$ as individual boxes, and compute residuals, as well as obtain guarantees in expectation, for individual boxes. 
However, this approach presents a challenge in defining the nonconformity score because the model only provides a set of predicted boxes. To address this, a pairing between predicted and ground truth boxes is necessary, which is commonly done using the Hungarian matching algorithm in the object detection literature. Since the nonconformity scores, and consequently the conformal guarantee, depend on this pairing, we consider it to be an integral part of the conformalization procedure. This operation is typically performed using a specific criterion.

considering a threshold on the \textit{IoU} (Intersection over Union) score of boxes: a prediction may be considered matched if it has a sufficient score. The spectrum of predictions the guarantee applies to, as well as the size of margins depends strongly on this threshold. It is most important to note that this guarantee will therefore apply exclusively to true positives. 

We further define multiple nonconformity scores, per coordinate, or a unique one per box, be it additive or multiplicative. We follow \citet{degrancey_2022_detection} and the generic Split \cp presented in Algorithm~\ref{alg:vanilla-split-cp} with the addition of the previously mentioned matching rule.

\subsubsection{Nonconformity scores for object detection} \label{sec:cp-od-scores}
Let $k=1, \dots, n_{box}$ index every ground-truth box in $\dcal$ that was detected by $\f$, disregarding their source image.  
Let $Y^{k} = (x_{\text{min}}^{k}, y_{\text{min}}^{k}, x_{\text{max}}^{k}, y_{\text{max}}^{k})$ be the coordinates of the $k$-th box and $\widehat{Y}^{k} = (\hat{x}_{\text{min}}^{k}, \hat{y}_{\text{min}}^{k}, \hat{x}_{\text{max}}^{k}, \hat{y}_{\text{max}}^{k})$ its prediction.

The nonconformity score, which we refer to as \textbf{additive}, is defined as: 
\begin{equation} \small
    R_{k} = \Big( 
     \hat{x}_{\text{min}}^{k} -     {x}_{\text{min}}^{k}, \,
     \hat{y}_{\text{min}}^{k} -     {y}_{\text{min}}^{k},\,
         {x}_{\text{max}}^{k} - \hat{x}_{\text{max}}^{k},\,
         {y}_{\text{max}}^{k} - \hat{y}_{\text{max}}^{k}
         \Big) \quad \text{(additive score)}.
     \label{eq:score-additive}
\end{equation}

\noindent Also hinted by \citet{degrancey_2022_detection}, a \textbf{multiplicative} score can be defined:
\begin{equation} \small
    R_{k} = \Big(
          \frac{\hat{x}_{\text{min}}^{k} -     {x}_{\text{min}}^{k}}{\dx} ,
          \frac{\hat{y}_{\text{min}}^{k} -     {y}_{\text{min}}^{k}}{\dy} ,
           \frac{    {x}_{\text{max}}^{k} - \hat{x}_{\text{max}}^{k}}{\dx}, 
           \frac{    {y}_{\text{max}}^{k} - \hat{y}_{\text{max}}^{k}}{\dy}
          \Big) \quad \text{(multiplicative score)},
    \label{eq:score-multiplicative}
\end{equation}

\noindent where the prediction errors are scaled by the predicted width $\dx$ and height $\dy$. 

Since they aim to capture a bounding box, which is represented by four coordinates, this yields a multidimensional response.
This case is similar to multiple hypothesis testing, and to guarantee the coverage of the whole box, that is, the four coordinates at the same time, it is necessary to apply a statistical adjustment to the risk level $\alpha$; and they opt for a Bonferroni correction where, for each coordinate, the prediction set is built with a miscoverage rate of $\frac{\alpha}{4}$ (see Eq.~\ref{eq:add-pred-set} \&~\ref{eq:mul-pred-set}).

It is well-known \citep{Bland_1995_multiple} that the Bonferroni correction can be overly conservative.
To counter this, we propose and test the \textbf{max-additive} and \textbf{max-multiplicative} nonconformity scores, which are defined respectively as:

\begin{align}  \small
    R_{k}^{\text{max}} &= \text{max} \big\{ %\Big( 
     \hat{x}_{\text{min}}^{k} -     {x}_{\text{min}}^{k}, \,
     \hat{y}_{\text{min}}^{k} -     {y}_{\text{min}}^{k}, \,
         {x}_{\text{max}}^{k} - \hat{x}_{\text{max}}^{k}, \,
         {y}_{\text{max}}^{k} - \hat{y}_{\text{max}}^{k}
         \big\}  &(\text{max-additive}), 
     \label{eq:max-additive-score} \\
    R_{k}^{\text{max}} &= \text{max} \big\{
          \frac{\hat{x}_{\text{min}}^{k} -     {x}_{\text{min}}^{k}}{\dx} ,
          \frac{\hat{y}_{\text{min}}^{k} -     {y}_{\text{min}}^{k}}{\dy} ,
           \frac{    {x}_{\text{max}}^{k} - \hat{x}_{\text{max}}^{k}}{\dx}, 
           \frac{    {y}_{\text{max}}^{k} - \hat{y}_{\text{max}}^{k}}{\dy}
          \big\}   &(\text{max-multip.}). 
    \label{eq:max-multiplicative-score}
\end{align}  

\begin{figure}[t]
\vspace{-10mm}
\center
\includegraphics[width=0.99\textwidth]{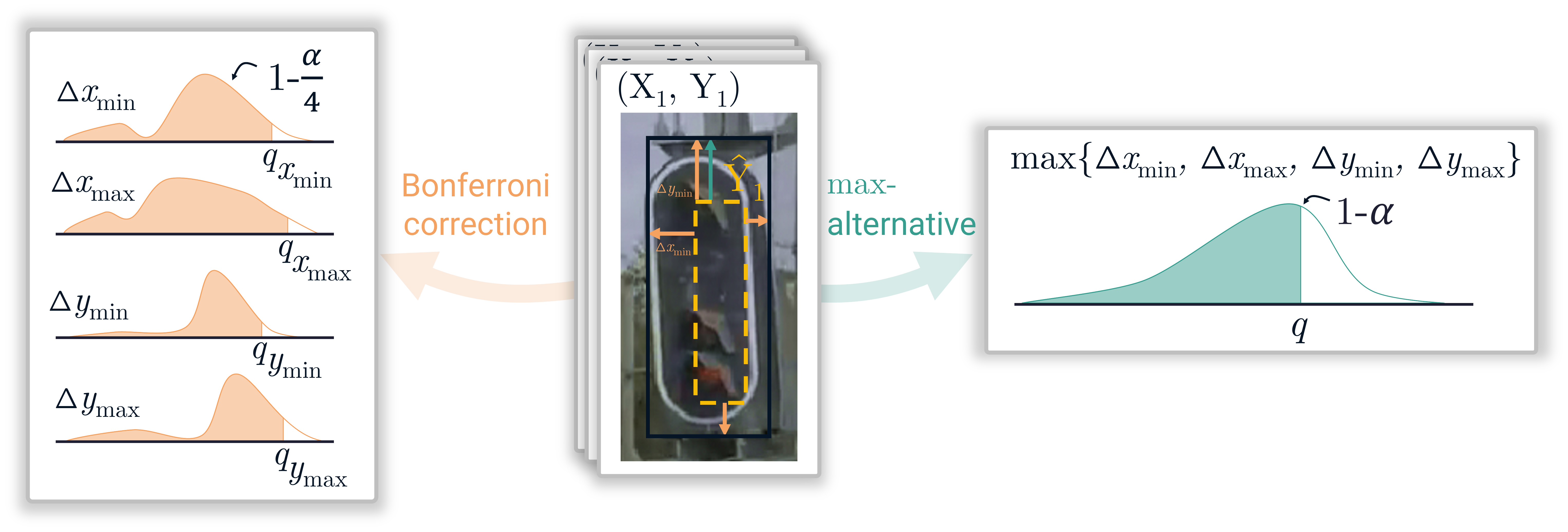}
\caption{\textbf{
Comparison of the previous Bonferroni approach vs our max-nonconformity score.}
 As explained in Section~\ref{sec:cp-od-scores}, we propose an alternative to the previously used Bonferroni correction which is overly conservative - as illustrated on the left, where four quantiles at the level $\frac{\alpha}{4}$ need to be estimated -- one per coordinate of the bounding box. The alternative - illustrated on the right - consists in calculating the quantile at the level $\alpha$ of the distribution of coordinate-wise maximum residuals.
 }
\label{fig:bonf-vs-max}
\vspace{-5mm}
\end{figure}

Both approaches are explained and compared on Figure~\ref{fig:bonf-vs-max}.
In general, it is not possible to determine a priori whether the additive or multiplicative score should be used.
For instance, \citet{degrancey_2022_detection} have operational reasons to prefer the additive margin: they are detecting pedestrians from the point of view of vehicles;
when objects are close to the camera, they have larger bounding boxes and multiplicative conformalization would yield margins that are too large to be operationally useful.

\subsubsection{Computing the conformal quantile}

After the nonconformity scores, the next quantity to compute is the conformal quantile.
For the case of additive and multiplicative scores with Bonferroni correction, we do:

\begin{equation} \small
    q_{1 - \frac{\alpha}{4}}^{c} = \lceil (n_{box}+1)(1-\frac{\alpha}{4}) \rceil\text{-th element of the sorted } \bar{R}^{c}, \forall c \in \{x_\text{min},y_\text{min},x_\text{max},y_\text{max}\}. 
\end{equation}

For the case of max-additive and max-multiplicative scores, the quantile is given by:
\begin{equation} \small
    q_{1 - {\alpha}}^{} = \lceil  (n_{box}+1)(1-\alpha) \rceil\text{-th element of the sorted } \bar{R}^{\text{max}}
\end{equation}

\subsubsection{Computing the prediction set} 
During inference, for a new observation $\xnew$, the \textbf{coordinates} of the {additive} and the {multiplicative} split conformal prediction boxes are given by:
\begin{align} \small
    \widehat{C}_{\alpha}(\xnew) = \big\{ 
        &\widehat{x}_{\text{min}} - q_{1-\frac{\alpha}{4}}^{x_\text{min}},\,
        \widehat{y}_{\text{min}} - q_{1-\frac{\alpha}{4}}^{y_\text{min}}, \, 
        \widehat{x}_{\text{max}} + q_{1-\frac{\alpha}{4}}^{x_\text{max}}, \,
        \widehat{y}_{\text{max}} + q_{1-\frac{\alpha}{4}}^{y_\text{max}} 
        \big\}, &(\text{additive)} 
    \label{eq:add-pred-set} \\ 
    \widehat{C}_{\alpha}(\xnew) = \big\{
        &\widehat{x}_{\text{min}} - \w \cdot q_{1-\frac{\alpha}{4}}^{x_\text{min}}, \,
        \widehat{y}_{\text{min}} - \h \cdot q_{1-\frac{\alpha}{4}}^{y_\text{min}}, \nonumber \\  %\,
        &\widehat{x}_{\text{max}} + \w \cdot q_{1-\frac{\alpha}{4}}^{x_\text{max}}, \,
        \widehat{y}_{\text{max}} + \h \cdot q_{1-\frac{\alpha}{4}}^{y_\text{max}} 
        \big\}. &(\text{multiplicative)}
    \label{eq:mul-pred-set} 
\end{align}

For the \textbf{max-error} conformalized versions, we get respectively:

\begin{align} \small
    \widehat{C}_{\alpha}(\xnew) = \Big\{\,
        &\widehat{x}_{\text{min}} - q_{1-\alpha},\,
        \widehat{y}_{\text{min}} - q_{1-\alpha},\, \nonumber \\
        &\widehat{x}_{\text{max}} + q_{1-\alpha},\,
        \widehat{y}_{\text{max}} + q_{1-\alpha}  
        \,\Big\}. & (\text{max-additive})
     \label{eq:max-add-box} \\
    \widehat{C}_{\alpha}(\xnew) = \Big\{ \,
        &\widehat{x}_{\text{min}} - \w \cdot q_{1-\alpha},\,
        \widehat{y}_{\text{min}} - \h \cdot q_{1-\alpha},\,  \nonumber \\
        &\widehat{x}_{\text{max}} + \w \cdot q_{1-\alpha},\,
        \widehat{y}_{\text{max}} + \h \cdot q_{1-\alpha}  
        \,\Big\}. & (\text{max-multiplicative box})
     \label{eq:max-mul-box}
\end{align}   

\subsection{Image-wise Conformalization and Conformal Risk Control}
\label{sec:crc}
\label{sec:image-wise}

Image-wise conformalization, as opposed to box-wise, considers our ground truth $Y_i$ to be defined as a set of bounding boxes. 
We therefore measure a nonconformity score at the image level and do not need a pairing algorithm. 
However, here we focus on avoiding false negatives, and do not consider the false positive rate in any of the following methods, and it does not affect the process of conformal prediction or risk control.

We compare here two families of approaches. The first is inherited from \citet{degrancey_2022_detection} and is rooted in a signed asymmetric Hausdorff distance. This method is therefore referred to as "Hausdorff" our experiments. The nonconformity score for this method is defined as the smallest margin such that, for a given image, a proportion $1-\beta$ of ground truth boxes is {\it entirely} covered by prediction boxes (even if it takes two prediction boxes that each cover half of a ground truth box). The parameter $\beta$ is set to 0.25 (arbitrarily) in the original work and we use the same value in our experiments.

This method presents a discontinuity which is common in \cp. Either the predictions cover over 75\% of ground truth boxes, or they don't. 
Thanks to 
\crc, this discontinuity has been removed and we can directly control the risk itself as the proportion of ground truth boxes that isn't covered.
We also further explore an evolution of this approach that instead considers as risk the average area of ground truth boxes that isn't covered, which therefore increases the penalty of misdetecting large ground truth boxes, while reducing it for smaller ones. In order to formulate those losses, let us define some notations.

\subsubsection{Losses for Conformal Risk Control}
For all calibration samples $(X_i, Y_i)_{i=1, \dots, n_{\text{cal}}}$, 
let $Y_i \in \{\varnothing,  \R^{1\times 4},  \R^{2\times 4}, \dots \}$ be the set of ground-truth bounding boxes included in $X_i$, which could be empty. 

The {\bf box-wise recall} loss is defined as the proportion of boxes that is not entirely covered by prediction boxes.  It is formulated as:

\begin{equation}
  L_{i}^{OD-box}(\lambda) = \ell \big(\widehat{\C}_{\lambda}(X_{i}), Y_{i} \big) =
    \begin{cases}
      0 & \text{if } Y_{i} = \varnothing \frac{}{} \\
      1 - \frac{1}{n_{i}}\sum_{k} \mathds{1}_{Y_{i}^{k}  \subseteq \bigcup_{j} \widehat{Y}_{i}^{j}}  & \text{otherwise}.
    \end{cases}       
    \label{eq:crc-loss-od-box}
\end{equation}
This formulation of the loss implies that multiple prediction boxes can be used to cover a single object and be considered correct. There is no discrimination between a single prediction box covering the whole ground truth box, and hundreds of pixels-sized boxes covering the ground truth too.
On the other hand, the pixel-wise recall is designed to tolerate partly covered ground truth, and is smoother than the box-wise recall. It requires further definition as follows.

Let $\mathcal{A}(Y_i^k) $ be the area (in terms of pixels) covered by the box $Y_i^k$. Moreover, let $Y_i^k \cap \bigcup_{j} \widehat{Y}_{i}^{j}$ denote the area of the intersection of the ground truth box with all predicted boxes.
The {\bf pixel-wise recall} loss is then defined as the average proportion of the area of ground truth boxes that isn't covered by predicted boxes. It is formulated as:
 
\begin{equation}
  L_{i}^{OD-pixel}(\lambda) = \ell \big(\widehat{\C}_{\lambda}(X_{i}), Y_{i} \big) =
    \begin{cases}
      0 & \text{if } Y_i = \varnothing \\
      1 -  \frac{1}{n_{i}}\sum_{k}\frac{ \mathcal{A} \big(Y_{i}^{k} \cap \bigcup\limits^{j} \widehat{Y}_{i}^{j} \big) }{ \mathcal{A} \big(Y_{i}^{k} \big) } & \text{otherwise}.
    \end{cases}       
    \label{eq:crc-loss-od-px}
\end{equation}

This loss, as the previous ones, tolerates multiple boxes used to cover a single ground truth, even partly in this case.
Moreover, this loss is expected to be impacted more by larger ground truth boxes, as models tend to very rarely predict boxes that are too small for small ground truths. It also can be seen as a further relaxation (smoothing) of the previous loss.

\subsubsection{Computation of Prediction Sets}
With these elements, we can apply Conformalized Risk Control to a pre-trained \od predictor $\f$, provided that we have access to some calibration data.
The details are in Algorithm~\ref{alg:crc-od}.

\begin{algorithm2e} 
\caption{(image-wise) conformally risk-controlled \od: \textit{conformalization}}% -- coordinate-wise}
\label{alg:crc-od}
 \DontPrintSemicolon 
    \begin{enumerate} \itemsep0em
    \item Split (disjointly) training data: $D_{\text{train}} = D_{\text{fit}} \cupdot D_{\text{cal}}$ 
    \item Fit (or fine-tune) the predictor $\widehat{f}$ on $D_{\text{fit}}$
    \item Compute the losses $L_i^{OD}$ on $D_{\text{cal}}$
    \item Estimate $\widehat{\lambda}$ as in Equation~\ref{eq:crc-lambda-estimation}
    \end{enumerate}
\end{algorithm2e}

During inference, we build the prediction set identically (although with $\lambda$ replacing quantiles) as in the box-wise method, for all three conformal methods:

\begin{equation} 
    \widehat{C}_{\widehat{\lambda}}(\xnew) = \Big\{  \,
        \widehat{x}_{\text{min}} - \widehat{\lambda} ,\,
        \widehat{y}_{\text{min}} - \widehat{\lambda},\, 
        \widehat{x}_{\text{max}} + \widehat{\lambda},\,
        \widehat{y}_{\text{max}} + \widehat{\lambda}
        \, \Big\}. \quad (\text{CRC additive)}
    \label{eq:crc-add-pred-set}
\end{equation}

\begin{equation} 
    \widehat{C}_{\widehat{\lambda}}(\xnew) = \Big\{  \,
        \widehat{x}_{\text{min}} - \w \cdot \widehat{\lambda} ,\,
        \widehat{y}_{\text{min}} - \h \cdot \widehat{\lambda},\, 
        \widehat{x}_{\text{max}} + \w \cdot \widehat{\lambda},\,
        \widehat{y}_{\text{max}} + \h \cdot \widehat{\lambda}
        \, \Big\}. \quad (\text{CRC multiplicative)}
    \label{eq:crc-mult-pred-set}
\end{equation}

\section{Experiments} \label{sec:experiments}

This section elaborates on the experiments performed on our created dataset. Initially, we describe the experimental setup, including the models utilized, fine-tuning process, and conformal parameters. Subsequently, we analyze the experimental outcomes based on two primary metrics: Stretch and Empirical coverage.

\begin{figure}[tbh]
%\label{fig:boxwise}
\floatconts
  {fig:boxwise}
  {\caption{{\bf Box-wise conformalization on an image with traffic signals at different scales.}
    Bounding boxes as predicted by the DiffusionDet predictor in \textcolor{yellow}{yellow}, conformalized boxes in \textcolor{purple}{purple} and ground truth in \textcolor{blue}{blue}. Cropped for readability.}}
  {%
    \subfigure[max-additive]{\label{fig:1cp-additive}%
      \includegraphics[width=0.4\linewidth]{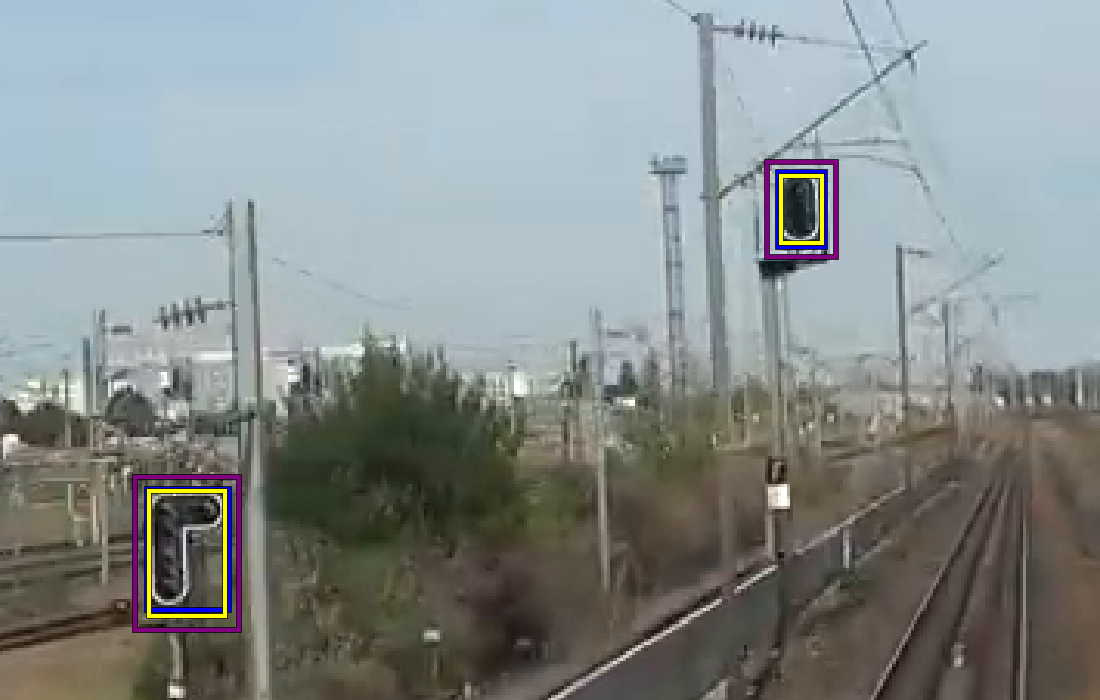}}%
    \qquad
    \subfigure[max-multiplicative]{\label{fig:1cp-multiplicative}%
      \includegraphics[width=0.4\linewidth]{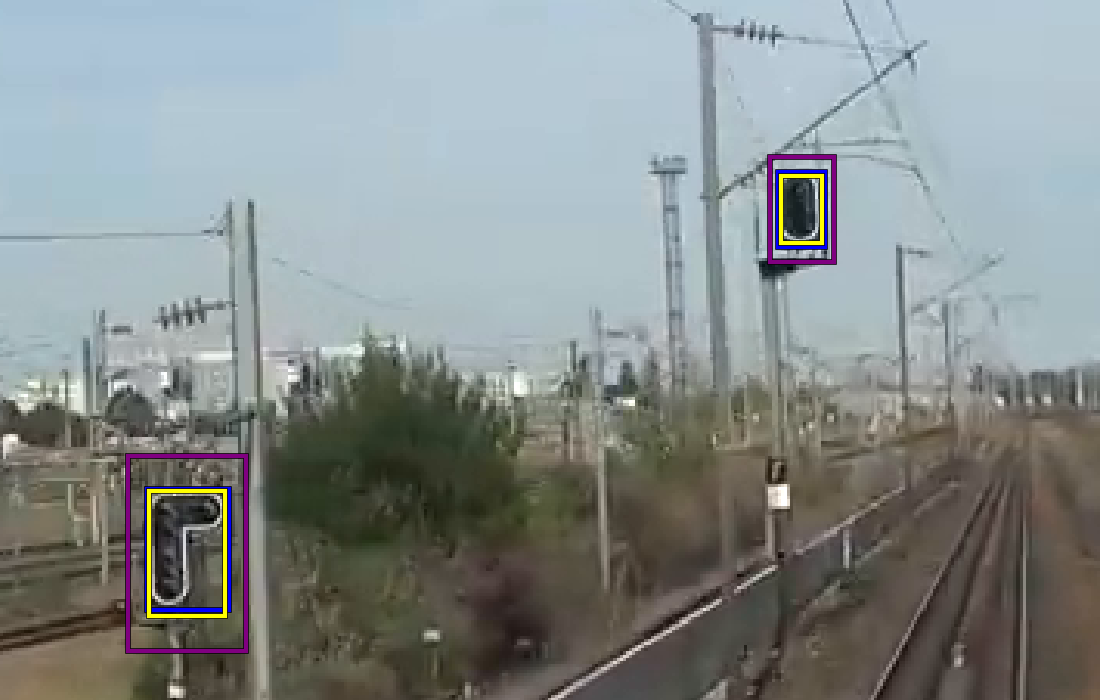}}
    \subfigure[additive]{\label{fig:4cp-additive}%
      \includegraphics[width=0.4\linewidth]{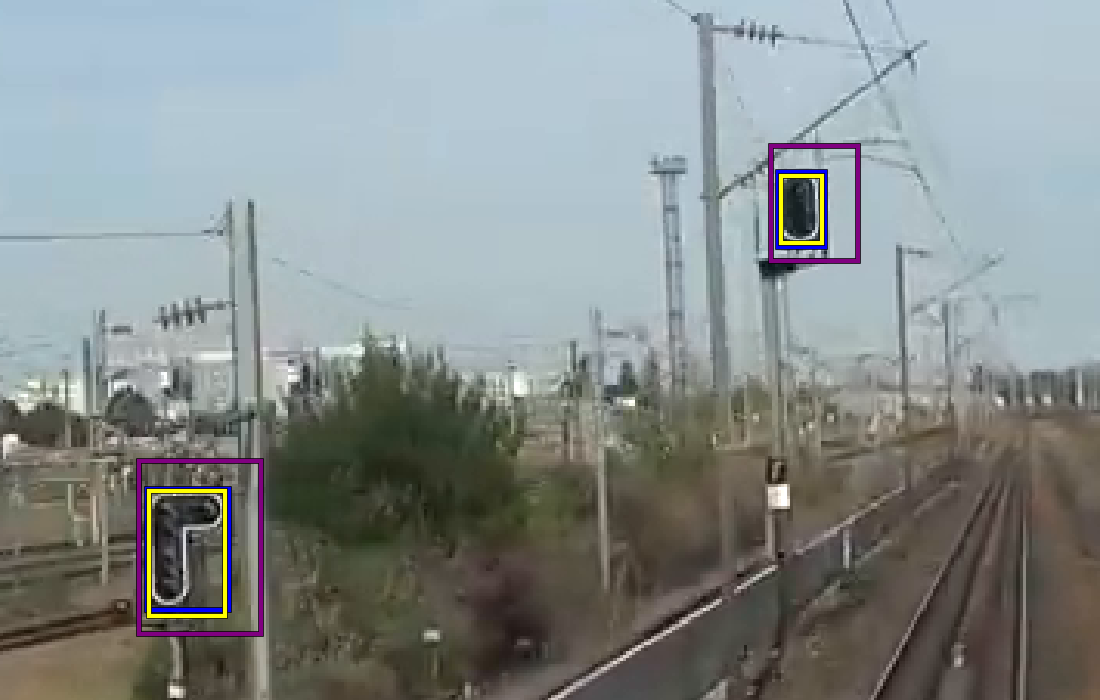}}%
    \qquad
    \subfigure[multiplicative]{\label{fig:4cp-multiplicative}%
      \includegraphics[width=0.4\linewidth]{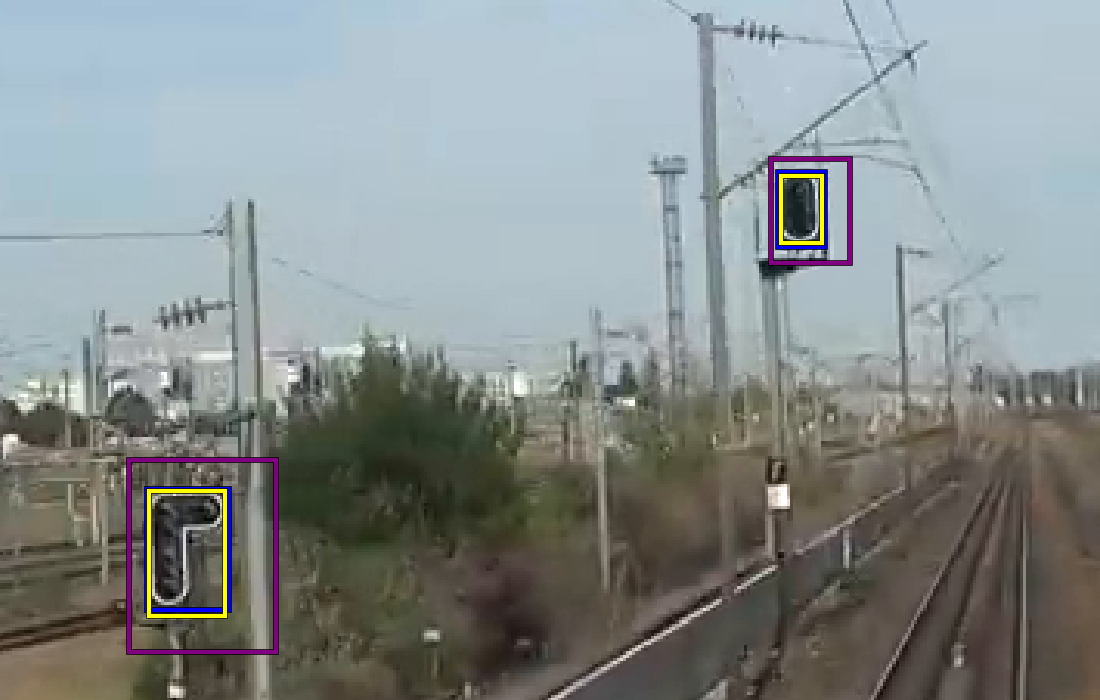}}
      \vspace{-5mm}
  }
\end{figure}

\subsection{Setup of experiments} 
We consider 2 settings for our experiments. One with two data splits for the three selected pretrained \od models, and another with an additional split for fine-tuning these pretrained models. We then apply run the conformal procedure on the calibration set, and evaluate results based on metrics.

In the literature \citep{Sesia_2020_comparison} we found that between 10\% to 50\% of $\dtrain$ are set aside for $\dcal$ (no pretrained $\f$).
For our experiments with pretrained predictors, we split the 3414 data points into 1914 for calibration and 1500 for testing.
This resulted in a total of 1974 bounding boxes in $\dtest$.
For the other set of experiments on fine-tuned predictors, we set aside 1414 points to $\dfit$, 1000 to $\dcal$ and 1000 to $\dtest$, for a total of 1022 bounding boxes in $\dtest$.
Throughout the tests, we set $\alpha = 0.1$, the objectness threshold of the predictor to $0.3$ and the IoU threshold to $0.3$.
\subsubsection{Fine-tuning the detectors} \label{sec:fine-tuning}
As mentioned above, for the second batch of experiments we set aside a partition of data $\dfit$ to fine-tune \yolov{m} and \detr. We also attempted fine-tuning of the \ddet model but were eventually unsuccessful.

The fine-tuning on the \yolov{m} model was done with its standard learning procedure, on all layers, for 100 epochs with a learning rate of $0.001$. For the \detr model, only 3 epochs of fine-tuning were conducted, optimizing only the 10 top layers of the backbone. The Adam optimizer was used with learning rate $3\times10^{-5}$ and weight decay $10^{-6}$. A small amount of data augmentation was added, with brightness, contrast and saturation jittering at level 0.1. 
\subsubsection{Performance of baseline predictors}
\begin{table} [h] 
    \centering 
    \small
    \begin{tabular}{ ll  c }
    % &  & Average Precision \\
    %     \cmidrule(lr){2-3}
     & \ccol{Model} &  Average Precision \\
    \midrule
    Pretrained & YOLOv5m         & 0.23 \\
    & \detr    &  0.29 \\
    & \ddet     & \textbf{0.45} \\
    \midrule
    Finetuned & YOLOv5m & 0.36 \\
    & \detr & \textbf{0.42}\\
    \end{tabular}
    \caption{Comparing models via Average Precision for an IoU threshold $\geq 0.3$.}
    \label{tab:new-detector-metrics} 
\end{table}

We report in Table \ref{tab:new-detector-metrics} the performance in terms of average precision of the multiple models that we chose to use in our experiments, both in their pretrained and finetuned version, as a reference for the evaluation of the different models in terms of conformal prediction performance. We recall that the average precision metric is computed as the area under of recall-precision curve, for recall and precision values computed at different objectness thresholds, i.e. minimum confidence of the model in that its own predicted boxes contain indeed an object.

We notice that the pretrained DiffusionDet predictor outperforms all others, including fine-tuned ones, and as the conformal procedure is post-hoc and therefore based on the model's outputs, the obtained quantiles or margins should vary, potentially significantly between models.

\subsection{Evaluation metrics}

To assess and contrast the various conformalization techniques, we employ two types of metrics in our experiments. The first metric we introduce is called 'stretch', which computes the average ratio of the areas of the conformalized boxes to the area of their corresponding raw prediction boxes. It can be expressed more formally as:

\begin{equation}
    \label{eq:stretch}
    \textrm{Stretch}=\sum\limits_{i=1}^{n}\sum\limits_{j=1}^{n_{i}}\sqrt{\frac{\mathcal{A} \big(C(X_{i})^{j}\big)}{\mathcal{A} \big(\f(X_{i})^{j}\big)}}.
\end{equation}
This metric is reported respectively for box-wise and image-wise conformalization in  Table~\ref{tab:boxwise-cp-stretch} and Table~\ref{tab:imagewise-strech}. Moreover, this metric is expected to be biased towards the multiplicative method, as it measure itself a multiplicative coefficient of growth rather than an additive one. 

The second metric we use is the empirical coverage, or empirical risk. We compute the empirical coverage for \cp methods and empirical risk for \crc. These metrics are only useful to measure how close the test performance of the conformalized sets is to the desired one, and higher(or lower) does not imply better. In fact, a coverage significantly higher than the desired level implies conformalized boxes larger than what would have been needed for that application. The coverage is defined as :

\begin{equation}
    \label{eq:coverage}
    \sum\limits_{i} \mathds{1}_{Y_{i} \in \C_{\hat{\lambda}}(X_{i})} ,
\end{equation}
while the risk is characterized by:
\begin{equation}
    \label{eq:risk}
    \sum\limits_{i} \ell \big(\C_{\hat{\lambda}}(X_{i}), Y_{i} \big),
\end{equation}
for any loss defined in section \ref{sec:image-wise}.

\subsection{Results}
\label{sec:results}
\begin{figure}[!htb]
%\label{fig:imagewises}
\floatconts
  {fig:imagewises}
  {\caption{{\bf Image-wise conformalization on an image with several traffic signals (including true and false positives, and false negatives.}
    Bounding boxes as predicted by the DiffusionDet predictor in \textcolor{yellow}{yellow}, conformalized boxes in \textcolor{purple}{purple} and ground truth in \textcolor{blue}{blue}. Cropped for readability.}}
  {%
    \subfigure[Hausdorff addit.]{\label{fig:dg-additive}%
      \includegraphics[width=0.4\linewidth]{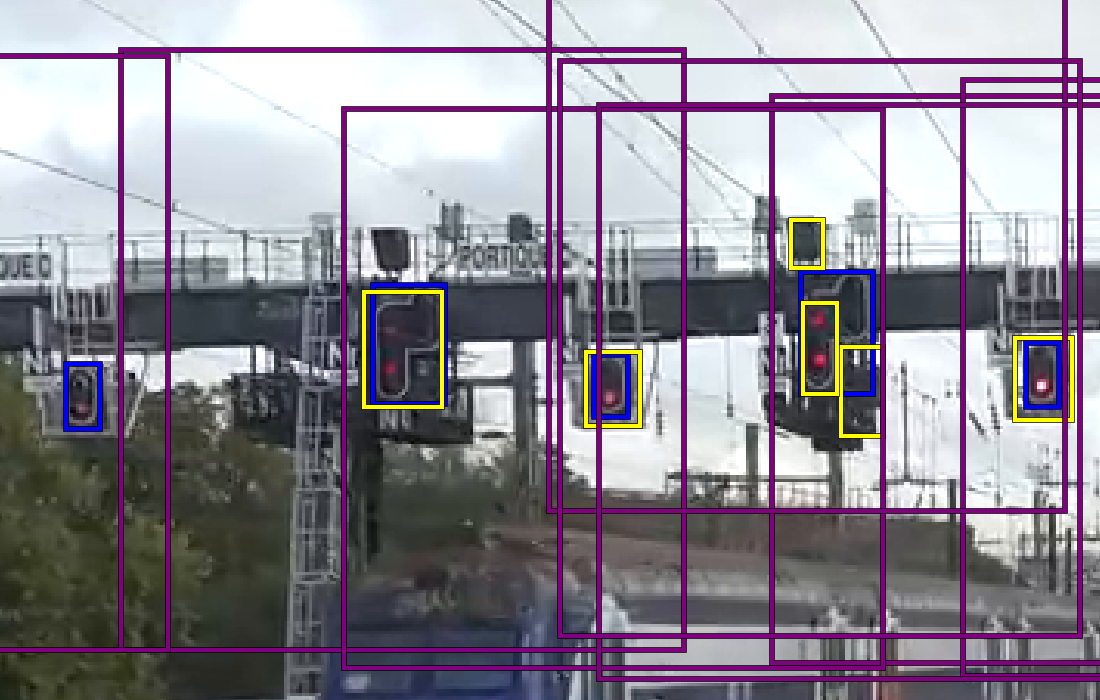}}%
    \qquad
    \subfigure[Hausdorff multip.]{\label{fig:dg-multiplicative}%
      \includegraphics[width=0.4\linewidth]{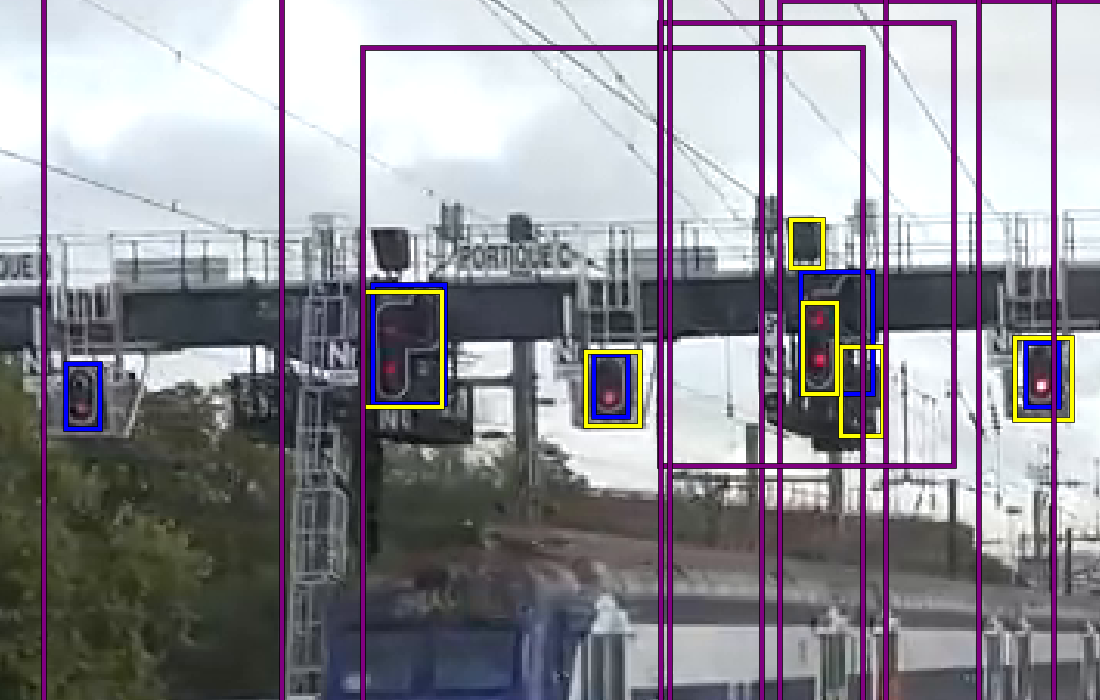}}
    \subfigure[Box Recall addit.]{\label{fig:box-additive}%
      \includegraphics[width=0.4\linewidth]{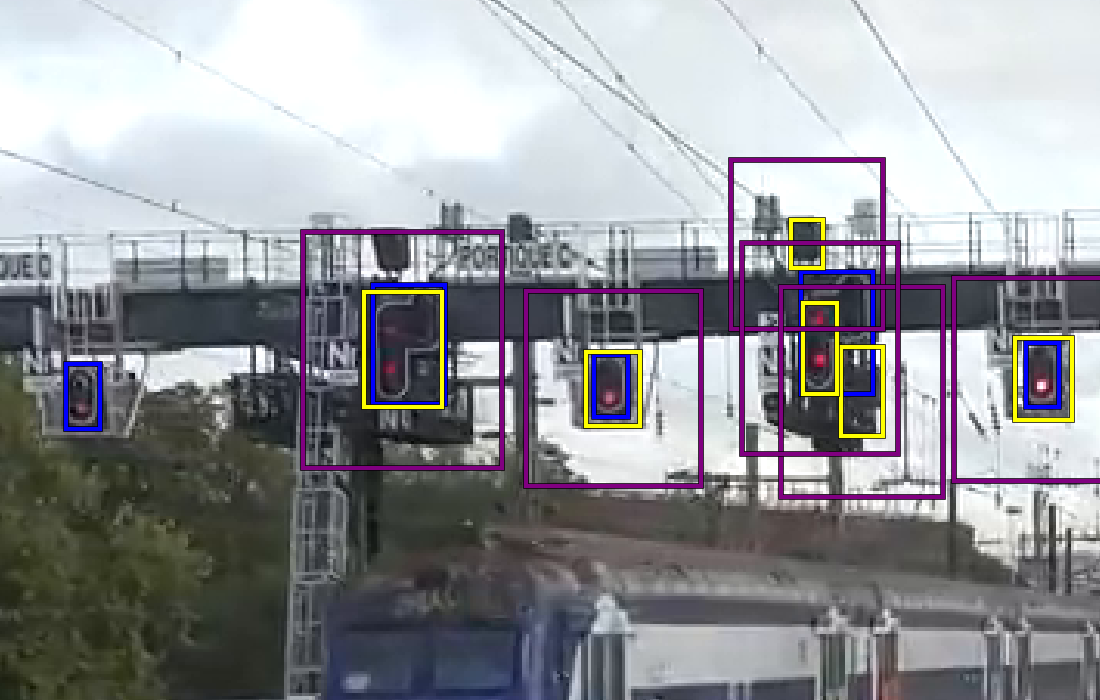}}
    \qquad
    \subfigure[Box Recall multip.]{\label{fig:box-multiplicative}%
      \includegraphics[width=0.4\linewidth]{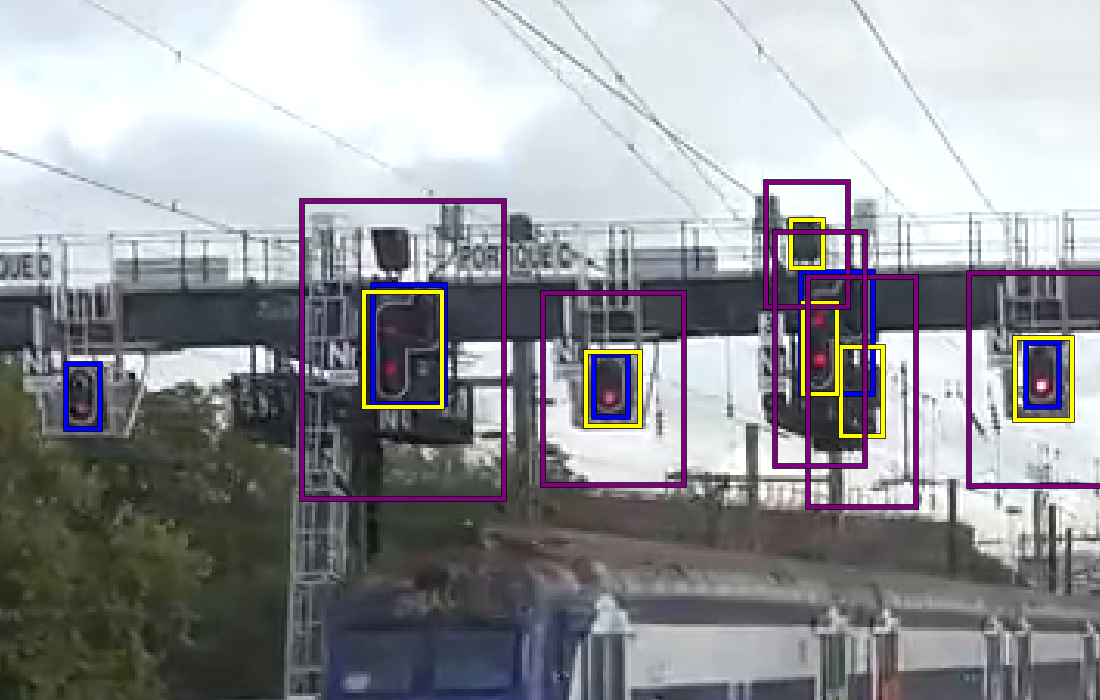}}
    \subfigure[Pixel Recall addit.]{\label{fig:px-additive}%
      \includegraphics[width=0.4\linewidth]{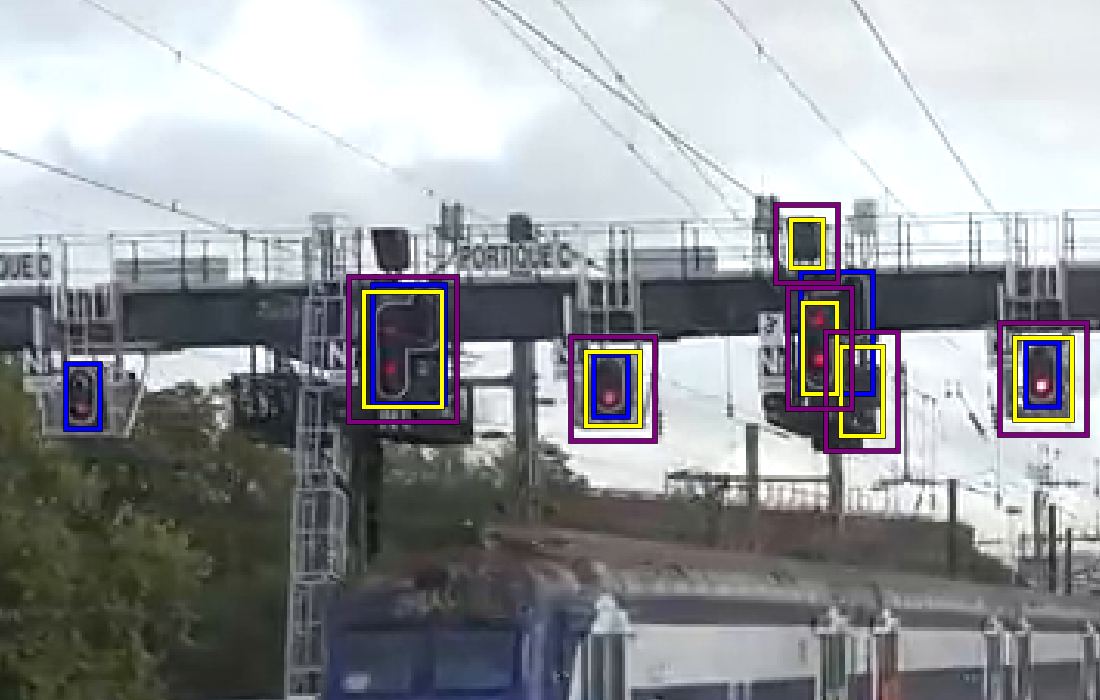}}%
    \qquad
    \subfigure[Pixel Recall multip.]{\label{fig:px-multiplicative}%
      \includegraphics[width=0.4\linewidth]{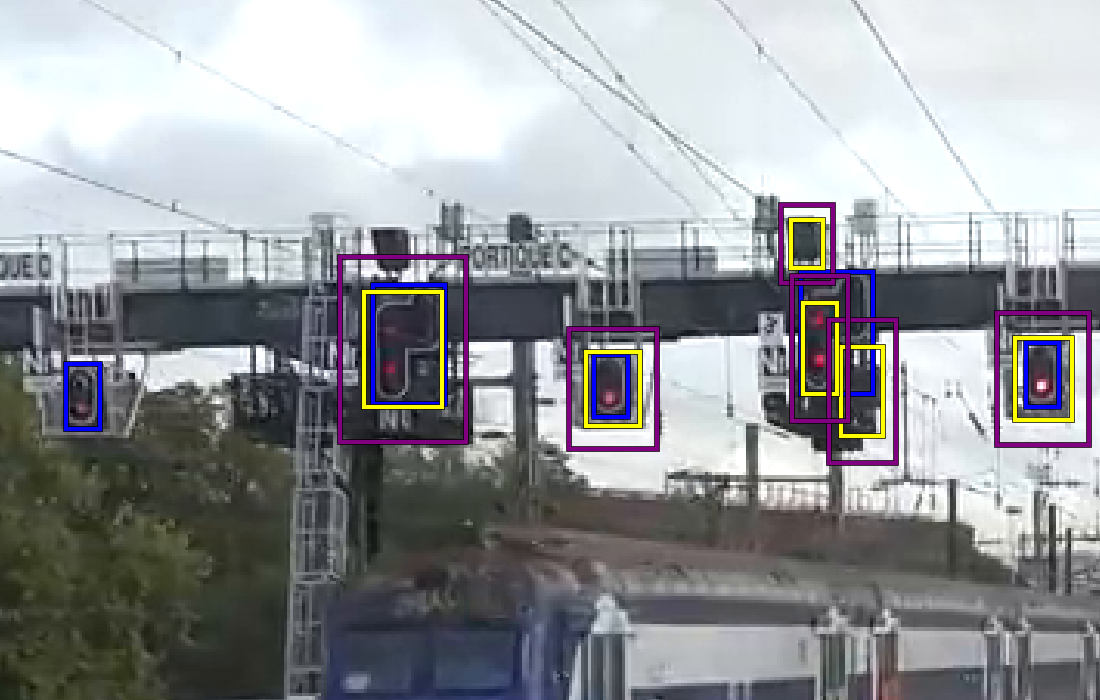}}
  }
\end{figure}

\subsubsection{Box-wise results}
We report in Table \ref{tab:boxwise-cp-stretch} and \ref{tab:boxwise-cp-coverage} respectively the stretch values and coverage of the multiple models and nonconformity scores we have experimented with.

In Table \ref{tab:boxwise-cp-stretch}, it appears that, model-wise, the YOLOv5m outperforms other methods (both pretrained and finetuned) as it leads to the smallest stretch in average. 
It is crucial to bear in mind that the box-wise conformalization method is exclusively applied to the true positives, which are the groud truth boxes that match with a prediction -- based on the IoU metric. Therefore, a model that generates fewer prediction may achieve a lower stretch value, provided that the predictions it generates are precise.

It is in fact the case as the YOLOv5m model, especially in its finetuned version outputs very few bounding boxes (of the right class) as compared to the DiffusionDet one.

\begin{table} [h] 
    \centering 
    \small
    \begin{tabular}{ ll  cccc}
    & \multicolumn{1}{c}{Model} & max-additive & max-multiplicative & additive & multiplicative \\
    \midrule
    Pretrained & YOLOv5m  & 1.35 & 1.45 & 1.45 & 1.56 \\
               & \detr             & 1.81 & 1.68 & 1.96 & 1.74 \\
               & \ddet             & 1.53 & 1.53 & 1.88 & 1.69 \\
    \midrule
    Finetuned & YOLOv5m & 1.33 & 1.35 & 1.34 & 1.36 \\
              & \detr             & 1.61 & 1.70 & 1.69 & 1.70 \\
    \end{tabular}
    \caption{\textbf{Box-wise}. Average stretch for multiple models and conformalization approaches.
    }
    \label{tab:boxwise-cp-stretch} 
\end{table}

In Table \ref{tab:boxwise-cp-coverage} model-wise we observe that the \yolov{m} has \textit{overall} the most calibrated coverage (close to the desired level $\alpha=0.1$). However, we note that generally the additive and multiplicative margins computed coordinate-wise seem to largely overcover, which is most likely due to the inner conservativeness of the Bonferroni correction. It appears that the DiffusionDet model with the max-additive or max-multiplicative nonconformity scores leads to the most calibrated prediction sets, close to the finetuned \yolov{m}.

In Figure~\ref{fig:boxwise} we can appreciate row-wise the effect of the distance (and therefore the size) of detection on the conformalized boxes under the multiple nonconformity scores. The closest sign (a),(c) is affected negatively, while the more distant one (b),(d) is affected positively (note that for (b), it is affected positively in terms of width but slightly negatively in terms of height, due to the height being larger than the width and being a factor in the scaling). Column-wise it appears clearly that the max approach has smaller margins than the Bonferroni approach, although it is noticeable that a bias from the model seems to appear, as there is a need for a larger correction in the top and right directions.

\begin{table} [h] 
    \centering
    \small
    \begin{tabular}{ ll  cccc }
     & \ccol{Model} & max-additive & max-multiplicative & additive & multiplicative \\
    \midrule
    Pretrained & YOLOv5m & 0.93 & 0.94 & 0.95 & 0.96 \\ 
               & \detr    & 0.99 & 0.98 & 1.00 & 0.99 \\
               & \ddet     & 0.92 & 0.92 & 0.96 & 0.96 \\
    \midrule
    Finetuned  & YOLOv5m& 0.94 & 0.92 & 0.95 & 0.93 \\
               & \detr    & 0.95 & 0.94 & 0.96 & 0.95 \\
    \end{tabular}
    \caption{\textbf{Box-wise}. Average coverage for multiple models and conformalization approaches.} 
    \label{tab:boxwise-cp-coverage} 
\end{table}

\subsubsection{Image-wise results}
We report in Table \ref{tab:imagewise-strech} and \ref{tab:imagewise-coverage} respectively the stretch values, and the coverage/risk values for the different models, correction types (additive or multiplicative) and guarantee type (conformal, pixel recall or box recall). In Table \ref{tab:imagewise-coverage} it is important to note that the coverage is reported for the Hausdorff \cp method, while the risk is reported for the \crc methods. As the \cp is a generalization of \cp, coverage can be considered a (non-smooth) loss, and in that case the risk would be 1 - coverage, as denoted in parenthesis in the table. We recall that better values are the ones close to the desired coverage or risk (resp. 0.9 and 0.1), and significantly better values than desired imply margins larger than necessary. Compared to the box-wise approach, the methods presented here produce well-calibrated prediction sets.

In both tables, we can notice that the two first lines are missing. This is due to the fact that there is no such value of $\lambda$ that solves eq. \ref{eq:crc-lambda-estimation}. In practice, this is due to a lack of predicted boxes (at the predefined objectness/confidence threshold). Even by increasing the size of predicted boxes infinitely, if there is no predictions for traffic lights by our model on an image, then for all $\lambda$, the loss associated with this image will be $1$. Therefore for the first two methods, is not possible to build a prediction set such that the guarantee at level $\alpha=0.1$ holds.

\begin{table} [h] 
    \centering 
    \small
    \begin{tabular}{ ll cc cc cc}
    & & \multicolumn{2}{c}{Hausdorff} & \multicolumn{2}{c}{Box Recall} & \multicolumn{2}{c}{Px Recall}\\
        \cmidrule(lr){3-4} \cmidrule(lr){5-6} \cmidrule(lr){7-8}
     & \ccol{Model} & addit. & multip. & addit. & multip. & addit. & multip. \\
    \midrule
    Pretrained & YOLOv5m &  ---  & --- & --- & --- & --- & --- \\
              & \detr    &  ---  & --- & --- & --- & --- & --- \\
              & \ddet    & 9.83 & 9.69 & 3.22 & 2.49 & 1.56 & 1.57\\
    \midrule
    Finetuned & YOLOv5m & 25.96 & 22.75 & 13.06 & 13.92 & 12.25 & 13.22\\
              & \detr &  7.74 & 10.00 & 3.36 & 2.75 & 2.16 & 2.05\\ 
    \end{tabular}
    \caption{\textbf{Image-wise}. Average stretch for multiple models and conformalization approaches. Missing values means an unattainable risk level, for our predictor and calibration data.}
    \label{tab:imagewise-strech} 
\end{table}

In Table \ref{tab:imagewise-strech}, we notice that the \yolov{m} that was best-performing in the box-wise experiments, is the worst performing here. This result is more aligned with expectations according to object detection performances presented in Table \ref{tab:new-detector-metrics}, and confirms that the performance of \yolov{m} in the previous task was due to the small number and accuracy of its predictions. Moreover, we observe that as the risk, or desired guarantee, is relaxed, the margins on the boxes decrease significantly. Therefore, while the obtained guarantees on the predicted boxes are slightly less strong with the box recall risk than the Hausdorff \cp, the conformalized boxes are significantly smaller, and even more so with the pixel recall risk.

On Figure~\ref{fig:imagewises}, appears more clearly than on the table the stretch of the different conformalization approaches. The Hausdorff approach leads to unusable bounding boxes in practice, while the others, in particular (e) lead to very reasonably sized bounding boxes, while holding a guarantee valid on images on average.

\begin{table} [h] 
    \centering 
    \small
    \begin{tabular}{ ll  cccccc}
    & & \multicolumn{2}{c}{Hausdorff} & \multicolumn{2}{c}{Box Recall} & \multicolumn{2}{c}{Px Recall}\\
        \cmidrule(lr){3-4} \cmidrule(lr){5-6} \cmidrule(lr){7-8}
     & \ccol{Model} & addit. & multip. & addit. & multip. & addit. & multip. \\
    \midrule
    Pretrained & YOLOv5m  &  ---  & --- & --- & --- & --- & --- \\
               & \detr             &  ---  & --- & --- & --- & --- & --- \\
               & \ddet      & (1-) 0.90 & (1-) 0.91 & 0.10 & 0.09 & 0.09 & 0.09 \\
    \midrule
    Finetuned  & YOLOv5m & (1-) 0.90 & (1-) 0.90 & 0.10 & 0.10 & 0.10 & 0.10 \\
               & \detr             & (1-) 0.87 & (1-) 0.87 & 0.12 & 0.12 & 0.12 & 0.12 \\ 
    \end{tabular} 
    \caption{\textbf{Image-wise}.  Average coverage and risk for multiple models and conformalization approaches. CRC 0.09: proportion of pixels missed by CRC-conformalized boxes. }
    \vspace{-3mm}
    \label{tab:imagewise-coverage} 
\end{table}

\subsection{Analysis}
A conformalized predictor can only reflect the quality (\eg accuracy) of its underlying base predictor $\f$.
If the latter misses many ground truth boxes, guaranteeing $(1 - \alpha) \, 100 \%$ correct predictions of a few boxes will still be a small number in the box-wise sense. Furthermore, in the image-wise sense, it will be impossible to reach the desired risk level with an underperforming model.
That is, conformalization is not a substitute for careful training or fine-tuning of a detection architecture, but a complementary tool to increase trustworthiness in the predictive models. 
For example, we can quantify \textit{how wrong} our predictions are, on average, based on the size of the conformal quantile $q_{1-\alpha}$: multiple predictors can be compared directly against our operational need (coverage, pixel recall, etc.). The interest of capturing the whole box can be operational: our \ml pipeline could rely on a conservative estimation of the ground truth to carry out a control operation (e.g. running a \ml subcomponent on the detection area).

Concerning the image-wise approach, we noticed very large variations between the different approaches, especially between the Hausdorff and the two others. This is due to a "threshold" effect: margins may be small at a certain level $\alpha$, but at a slightly lower level $\alpha-\epsilon$, one more ground truth box has to be included in order to satisfy the guarantee, and in the case the closest box non-covered box is distant, such as the leftmost on Figure~\ref{fig:imagewises}, the margin can dramatically explode. This effect is increased on the Hausdorff approach, as 75\% of boxes need to be covered on 90\% on images, while the box-wise approach requires 90\% of boxes to be covered in expectation, and therefore can largely fail on difficult images, and compensate on others.

\section{Conclusion} \label{sec:conclusion}

Given the insights from this investigation, we intend to develop an enhanced iteration of the dataset, which will serve as a dedicated and high-quality benchmark for evaluating conformal prediction in the domain of object detection, catering to both the scientific community and the transport industry.

It is noteworthy that conformal prediction operates under the assumption of exchangeable data. However, for the deployment of trustworthy AI components in the long run, the problem setting and underlying assumptions will need to be adapted to account for the dynamics of data streams to ensure reliable uncertainty quantification guarantees. This process will present theoretical and practical challenges concerning the construction and validation of datasets.

Our analysis revealed that the current success criterion for prediction relied on the complete coverage of the ground truth boxes which may be overly restrictive.
In practice, it may be adequate for a system to ensure coverage of a substantial portion of the ground truth. 
This direction deserves more interest from the industrial community, in order to reach much lower risk levels $\alpha$ for viable real-world applications.

Nevertheless, it is improbable that conformal prediction alone can achieve sufficiently low-risk levels. It will be imperative to study larger datasets employing cutting-edge models in tandem with custom-designed conformal methods, and collaborate with domain experts to chart a more definitive course towards certified object detection.

Lastly, it is worth noting that several concentration-based approaches have been developed and should be compared in depth to conformal ones.

\acks{
This work has benefited from the AI Interdisciplinary Institute ANITI, which is funded by the French “Investing for the Future – PIA3” program under the Grant agreement ANR-19-P3IA-0004.
The authors gratefully acknowledge the support of IRT Saint Exupéry and and the DEEL project.\footnote{\url{https://www.deel.ai}}}

%\clearpage
%\vskip 0.2in
\small
\bibliography{references}

\end{document}